\documentclass[letterpaper, 10 pt, conference]{ieeeconf}  

\IEEEoverridecommandlockouts                              
\usepackage{multirow}
\usepackage{cite}
\usepackage{amsmath,amssymb,amsfonts}
\usepackage{algorithmic}
\usepackage{graphicx}
\usepackage{textcomp}
\usepackage{xcolor}                                                          
\usepackage{authblk} 
\title{Accelerating Vision Transformers on Brain Processing Unit} 

\begin{document}

\author[1,2]{Jinchi Tang}
\author[1,2 *]{Yan Guo}
\affil[1]{Suzhou Institute for Advanced Research,USTC}
\affil[2]{University of Science and Technology of China}
\affil[*]{guoyan@ustc.edu.cn}

\maketitle
\thispagestyle{empty}
\pagestyle{empty}

\begin{abstract}

With the advancement of deep learning technologies, specialized neural processing hardware such as Brain Processing Units (BPUs) have emerged as dedicated platforms for CNN acceleration, offering optimized INT8 computation capabilities for convolutional operations. Meanwhile, Vision Transformer (ViT) models, such as the Data-efficient Image Transformer (DeiT), have demonstrated superior performance and play increasingly crucial roles in computer vision tasks.

However, due to the architectural mismatch between CNN-optimized hardware and Vision Transformer computation characteristics—namely, that linear layers in Transformers operate on three-dimensional data while BPU acceleration is designed for four-dimensional convolution operations—it is difficult or even impossible to leverage BPU's advantages when deploying Vision Transformers.

To address this challenge, we propose a novel approach that restructures the Vision Transformer by replacing linear layers and layer normalization operations with carefully designed convolutional operators. This enables DeiT to fully utilize the acceleration capabilities of BPUs, while allowing the original weight parameters to be inherited by the restructured models without retraining or fine-tuning.

To the best of our knowledge, this is the first successful deployment of Vision Transformers that fully leverages BPU acceleration. Extensive experiments on the ImageNet and flower classification datasets demonstrate the effectiveness of our approach. Specifically, the quantized DeiT-Base model achieves 80.4\% accuracy on ImageNet, compared to the original 81.8\%, while obtaining up to a 3.8$\times$ inference speedup. Our fine-tuned DeiT model on the flower classification dataset also achieves excellent performance, with only a 0.5\% accuracy drop for the DeiT-Base model, further demonstrating the effectiveness of our method.

\end{abstract}

\section{INTRODUCTION}

Recent years have witnessed the emergence of specialized embedded hardware devices, among which the Horizon Brain Processing Unit (BPU) stands out for its efficiency in handling deep learning workloads. The BPU, built on the ARM Cortex-A53 CPU core, implements a dedicated hardware design specifically optimized for quantized neural network operations. A distinctive feature of this architecture is its effective support for INT8 quantization in convolution operations, which significantly reduces memory bandwidth requirements while maintaining computational accuracy. The hardware encompasses specialized compute units that can efficiently process four-dimensional data streams, making it particularly well-suited for convolutional neural networks (CNNs). Through careful architectural optimizations, including dedicated datapaths and specialized compute units, the BPU achieves remarkable acceleration for quantized convolution operators.

On the other hand, Vision Transformer models~\cite{dosovitskiy2021image} have revolutionized computer vision tasks by introducing the self-attention mechanism. Among these models, the Data-efficient Image Transformer (DeiT)~\cite{touvron2021training} has shown particular promise by achieving competitive performance with much less training data and computational resources compared to its predecessors.

However, computation in Vision Transformer models differs significantly from traditional CNNs. Traditional CNN models, such as those based on MobileNet~\cite{howard2017mobilenetsefficientconvolutionalneural,sandler2019mobilenetv2invertedresidualslinear,howard2019searchingmobilenetv3}, primarily perform convolution operations on four-dimensional data streams with shapes like $[B, C, H, W]$, where $B$, $C$, $H$, and $W$ represent batch size, channel, height, and width, respectively. The BPU currently provides mature quantization and acceleration solutions for such operations, allowing for efficient deployment in various applications.

In contrast, Transformer-based models~\cite{vaswani2017attention} predominantly rely on linear layers, as employed in self-attention mechanisms and feed-forward networks (FFN). These architectural characteristics make it significantly challenging—if not impossible—to fully leverage the specialized features of the BPU platform, particularly its optimized convolution operations and four-dimensional tensor processing capabilities. As a result, when deploying DeiT vision transformers, only the CPU is available for transformer computation.

To address this challenge, we propose a novel approach by replacing the linear layers and layer normalization operations in the DeiT architecture with equivalent convolution operators. This architectural modification allows us to better utilize the BPU's hardware capabilities, particularly its optimized units for convolution processing and efficient INT8 quantization support. Importantly, the transformed model does not require retraining—the weights can be directly inherited from pre-trained DeiT model checkpoints. After quantization with Horizon's BPU algorithm toolchain, the modified model can be deployed on the BPU embedded platform, taking full advantage of its hardware-specific optimizations.

We conduct extensive experiments to verify the effectiveness of our approach on two datasets: ImageNet and a flower classification dataset. First, we test and compare the accuracy and efficiency of the optimized models and the original DeiT on ImageNet~\cite{deng2009imagenet}. Then, we fine-tune only the last layer of the optimized model on the flower classification dataset~\cite{2020mmclassification}. Our results show that the quantized models achieve minimal accuracy degradation across most model variants while attaining significant speedup, ranging from 1.3$\times$ to 3.8$\times$ compared to their floating-point counterparts. We also specifically test samples with known incorrect labels~\cite{northcutt2021pervasivelabelerrorstest} in the ImageNet validation set, and observe that our models successfully predict the correct labels.

In summary, our work makes the following contributions:
\begin{itemize}
    \item We propose a BPU-optimized Vision Transformer architecture and successfully develop deployable quantized DeiT models for embedded platforms using Horizon's BPU algorithm toolchain. To the best of our knowledge, this is the first successful attempt to deploy DeiT on BPU and utilize its acceleration capabilities.
    \item We conduct comprehensive experiments to validate our approach. On the ImageNet dataset~\cite{deng2009imagenet}, our quantized models achieve satisfactory accuracy and promising speedup compared to the original implementations.
    \item We further extend our evaluation to a flower classification dataset~\cite{2020mmclassification}, where we fine-tune DeiT models on this new dataset and analyze their performance before and after quantization, again achieving minimal accuracy loss.
\end{itemize}

The remainder of this paper is organized as follows: Section~\ref{sec:related_work} reviews related work, including DeiT and quantization methods. Section~\ref{sec:proposed_method} details our proposed approach. Section~\ref{sec:experiments} presents experiments on the two datasets. Section~\ref{sec:conclusion} concludes the paper.

\section{Related Work}
\label{sec:related_work}
Our research builds primarily upon two fields: Vision Transformers, particularly the DeiT architecture, and model quantization techniques.

\subsection{Vision Transformers and DeiT}
Transformers~\cite{vaswani2017attention}, initially designed for natural language processing tasks, have revolutionized the deep learning landscape since their introduction. Vision Transformer (ViT)~\cite{dosovitskiy2021image} successfully adapted this architecture to computer vision tasks by treating images as sequences of patches. Numerous subsequent works have further enhanced its capabilities. Among them, DeiT~\cite{touvron2021training} stands out for its outstanding efficiency and accuracy.

ViT originally requires pre-training on the JFT-300M~\cite{sun2017revisitingunreasonableeffectivenessdata} dataset followed by fine-tuning on ImageNet to surpass CNN performance, which significantly increases model training cost and complexity. To address this issue, DeiT drew inspiration from knowledge distillation~\cite{hinton2015distilling} and proposed a teacher-student training method. DeiT leverages a CNN model~\cite{abnar2020transferringinductivebiasesknowledge} as the teacher to guide training, making direct training on ImageNet possible without the need for large-scale pre-training data. 

Additionally, the authors of DeiT employ extensive efforts, such as comparing soft~\cite{wei2020circumventing} and hard distillation methods, applying data augmentation techniques~\cite{cubuk2018autoaugment,cubuk2019randaugment,zhong2020random} on ImageNet, and utilizing patch embedding to segment images into smaller patches for encoding. Through these approaches, DeiT ultimately achieves state-of-the-art (SOTA) performance with a relatively small number of parameters. Compared with ViT, the DeiT-Small version (22M parameters) achieves accuracy comparable to the ViT-Base version (86M parameters). Since BPU is equipped with limited hardware resources, we choose DeiT as our target ViT model and focus on its smaller variants.

\subsection{Model Quantization}
Quantization has become increasingly important for deploying deep learning models on resource-constrained devices. This technique reduces model precision from floating-point to lower-bit representations while striving to preserve model accuracy. Numerous works and related frameworks supporting quantized inference~\cite{nagel2021whitepaper,krishnamoorthi2018quantizing,jacob2017quantizationtrainingneuralnetworks,reed2022torchfxpracticalprogramcapture} have been proposed. Quantization methods are generally categorized into Post-Training Quantization (PTQ) and Quantization-Aware Training (QAT).

QAT~\cite{zhou2018dorefa,choi2018pact} introduces quantization operations into the training pipeline by simulating quantization effects during both the forward and backward passes. This approach enables the model to learn optimal quantization parameters during training, resulting in better accuracy compared to PTQ. However, it requires computationally expensive retraining and careful tuning of additional hyperparameters, including learning rate scheduling, quantization-aware loss functions, and bit-width configuration for different layers.

In contrast, Post-Training Quantization (PTQ) directly converts pre-trained models to lower precision by analyzing the distribution of model weights and activations without retraining, making it a more efficient and practical solution for deployment. Furthermore, PTQ enables layer-wise~\cite{hubara2020improving} quantization rather than QAT's network-wide approach, and requires only a small, unlabeled calibration dataset~\cite{hubara2021accurate} without the need for additional training. The calibration set is used to collect statistical information about activation distributions, which guides the determination of optimal quantization parameters for each layer. This significantly reduces the computational resources needed for quantization while improving deployment efficiency.

The BPU is a hardware device designed to accelerate quantized operators. Specifically, for convolution operations on tensors with shapes such as [B, C, H, W], Horizon's official toolchain can perform quantization preprocessing using algorithms such as the KL divergence-based method~\cite{kullback1951information}. Subsequently, during model inference, the BPU hardware enables accelerated computation of convolution operators at INT8 precision. However, the BPU's inherent acceleration does not support linear and LayerNorm computations, resulting in the inability to run inference for Vision Transformer models directly on the BPU. Our work addresses this limitation by reformulating these operators to be compatible with BPU hardware.

\section{Proposed BPU-Utilized DeiT Model}
\label{sec:proposed_method}
In this section, we analyze the structure of DeiT and present our proposed quantization scheme, enabling the model to be deployed on edge computing platforms such as BPU. We detail the analysis and design of the key components of the quantized model and introduce specific enhancements implemented to improve performance and efficiency.

\subsection{BPU-Optimized Operators}
As previously introduced, the BPU (Brain Processing Unit) is a processing unit specifically optimized for 2D convolution operations. To fully exploit this feature, it is necessary to redesign linear and LayerNorm operators in the Vision Transformer, adapting them to convolution operations. Such operators are primarily found in the multi-head self-attention mechanism and the feed-forward network. Therefore, to optimize the DeiT model for deployment on edge computing platforms, we focus on restructuring the linear and LayerNorm operators in all multi-head self-attention and FFN modules of DeiT.

For linear layers, we transform them into equivalent pointwise convolutions, where the weight matrix $\mathbf{W} \in \mathbb{R}^{O \times I}$ is reshaped to a 4D tensor $\mathbf{W'} \in \mathbb{R}^{O \times I \times 1 \times 1}$ through two consecutive unsqueeze operations. Similarly, for existing pointwise convolutions, the weight tensor is extended to the same 4D format. The input tensor is transformed to a consistent $(B, C, 1, T)$ format by transposing the channel dimension from $(B, N, C)$ to $(B, C, 1, N)$. This conversion enables efficient computation on the BPU while maintaining mathematical equivalence. To verify the effectiveness of the transformation, we compare the restructured model with the original. Across multiple inputs, the average relative tolerance between the outputs of the two models remains within $10^{-2}$ and absolute tolerance within $10^{-3}$.

Similarly, we adapt Layer Normalization (LayerNorm) operators for BPU execution through a novel reformulation using 4D convolution operations. The standard LayerNorm computation $y = \frac{x - \mathbb{E}[x]}{\sqrt{\text{Var}[x] + \epsilon}} \cdot \gamma + \beta$ is decomposed into a sequence of $1 \times 1$ convolution operations.

\begin{figure}[!h]
\centering
\includegraphics[width=0.48\textwidth, height=0.3\textheight, keepaspectratio]{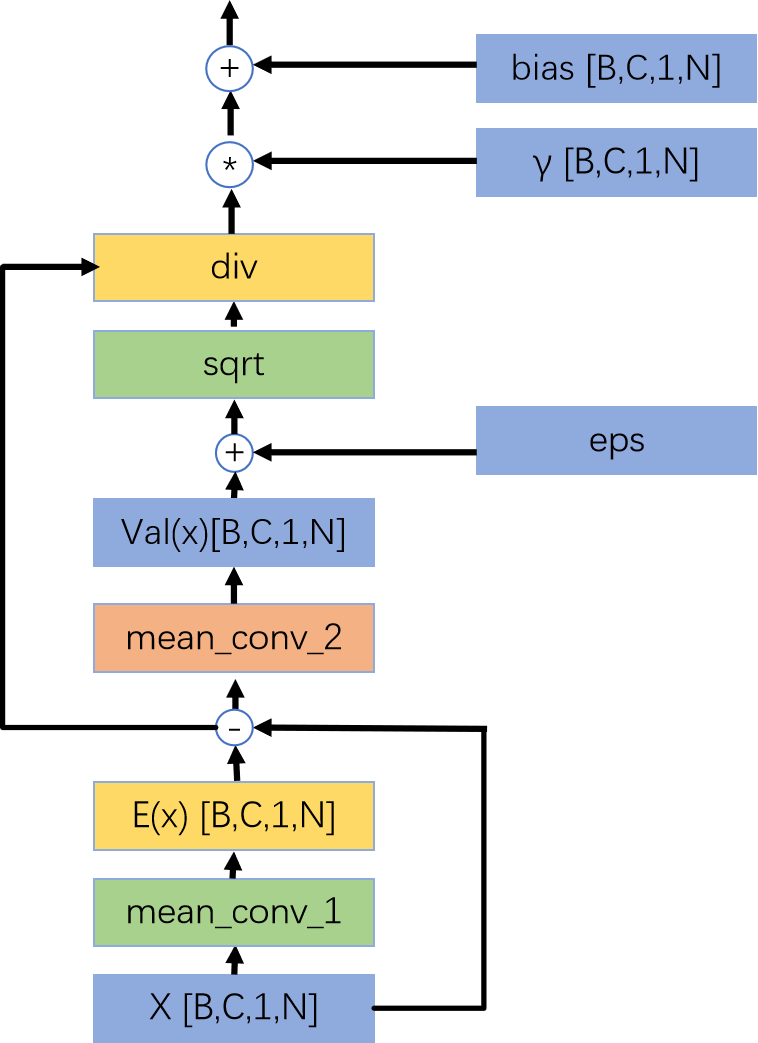}
\caption{Architecture of BPU-optimized LayerNorm implementation using $1 \times 1$ convolutions, illustrating the internal computational structure of LayerNorm on the BPU platform via convolution operations.}
\label{fig:layernorm}
\end{figure}

As shown in Fig.~\ref{fig:layernorm}, the BPU LayerNorm reformulates the normalization process into a series of convolution operations to leverage hardware acceleration. The average is computed using a $1 \times 1$ convolution ($\text{mean\_conv\_1}$) where all weights are set to $\frac{1}{H}$ ($H$ being the hidden dimension). The variance is then calculated by first subtracting the mean, squaring the result, and applying another $1 \times 1$ convolution ($\text{mean\_conv\_2}$). The $\gamma$ (weight) and $\beta$ (bias) parameters are reshaped to match the dimensions of the input tensor.

As with the linear operator reformulation, we evaluated the modified LayerNorm operator. The outputs of the original PyTorch LayerNorm and the BPU-adapted implementation on random input data are compared, with a relative tolerance of $10^{-2}$ and absolute tolerance of $10^{-3}$, fully demonstrating the effectiveness of our approach.

\subsection{Transformer Blocks on the BPU Platform}

\begin{figure}[!h]
  \centering
  \includegraphics[width=0.48\textwidth]{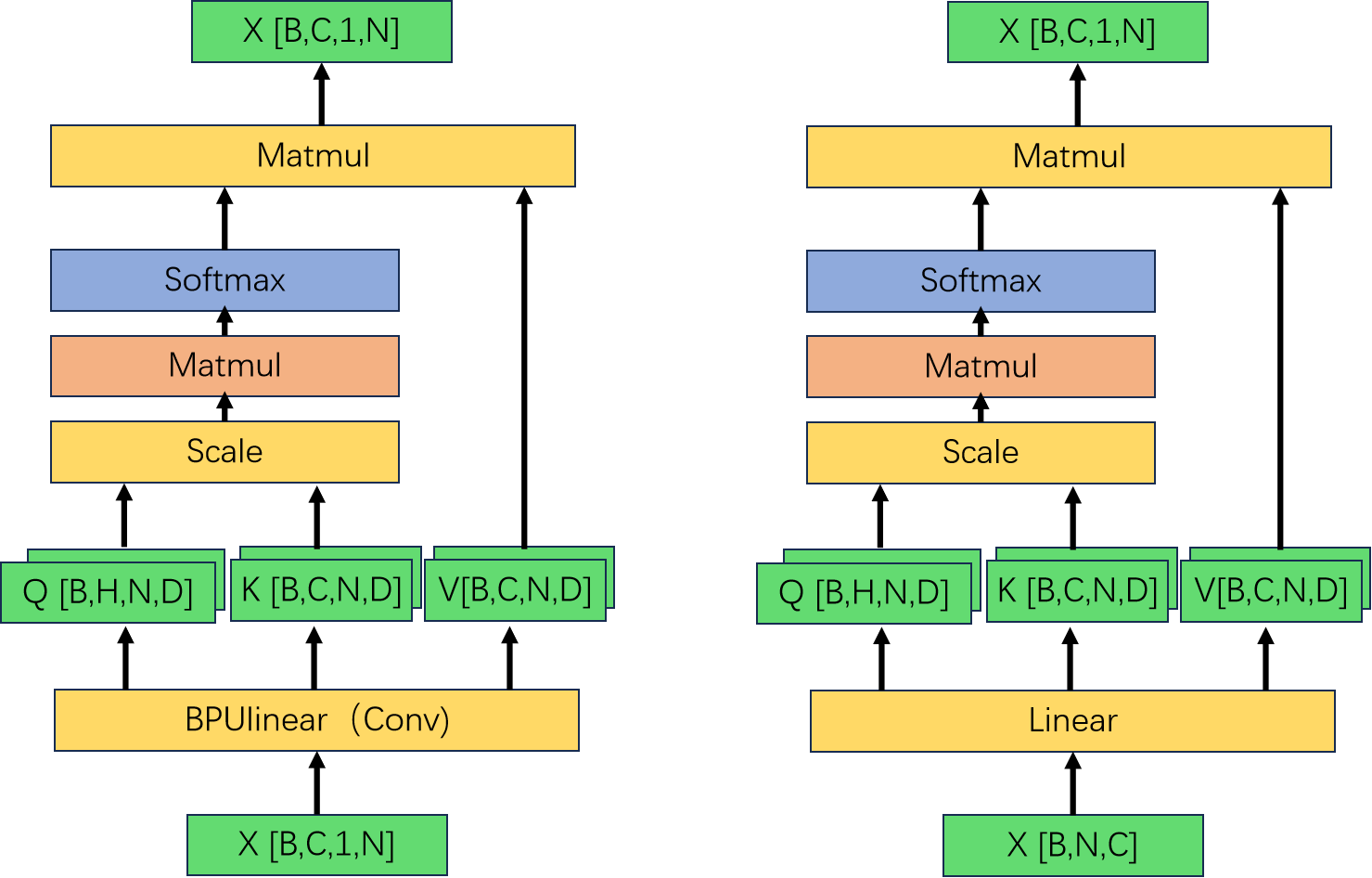}
  \caption{Comparison of attention mechanisms. (Left) The proposed BPU-optimized attention mechanism with hardware-friendly reformulation using convolution operations. (Right) The standard attention structure of the original implementation with matrix multiplications and softmax operations.}
  \label{fig:transformer_attention}
\end{figure}

\begin{figure}[!h]
  \centering
  \includegraphics[width=0.48\textwidth]{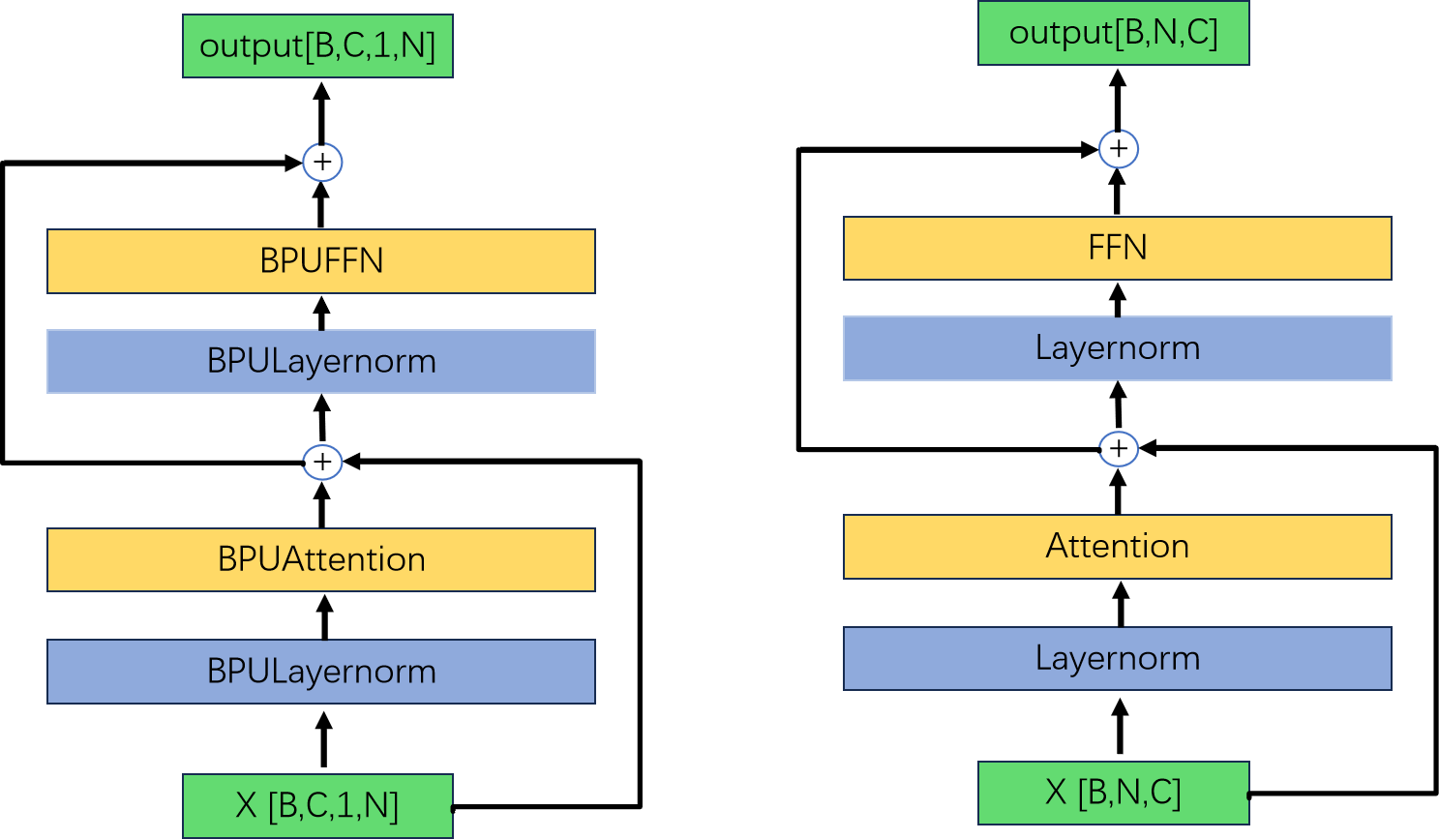}
  \caption{Comparison of Transformer block architectures. (Left) The BPU-optimized Transformer block with hardware-efficient reformulation for specialized acceleration. (Right) The standard Transformer block structure showing the conventional implementation with typical attention, MLP, LayerNorm, and residual connections.}
  \label{fig:transformer_bpu}
\end{figure}

With the previously established operator transformations as our foundation, we systematically adapt each component of the Transformer block to efficiently utilize the BPU's computational capabilities. The architecture of these optimized Transformer blocks is illustrated in Figure~\ref{fig:transformer_attention} and Figure~\ref{fig:transformer_bpu}.

As illustrated in Fig.~\ref{fig:transformer_attention}, the left side demonstrates the specialized attention mechanism optimized for the BPU platform, while the right side shows the conventional attention mechanism. In the BPU-optimized attention mechanism, the original Linear layers are replaced with 2D convolution layers. As described previously, this operator produces results nearly identical to the original linear operation, enabling us to incorporate convolution layers into the attention mechanism without loss of accuracy. Similarly, the FFN layers in the transformer follow the same strategy, integrating convolution operations while preserving output equivalence. The only difference in the BPU attention mechanism's output is tensor shape, which, after appropriate reshaping, maintains consistency with the original attention mechanism. This design allows us to leverage the BPU's acceleration capabilities for convolution operations—particularly its optimized INT8 convolution computation—significantly improving model inference efficiency.

The left side of Figure~\ref{fig:transformer_bpu} presents the overall architecture of the BPU-optimized Transformer block, while the right side shows the conventional Transformer block. Aside from input and output dimensions, the overall structure remains consistent with the standard Transformer block, with primary differences in the internal operator implementations. Since the majority of the model structure remains unchanged, and the convolutional operators differ only in dimensions, our optimized model can inherit the original model weights on the BPU platform without retraining or additional fine-tuning.

\section{Experiments}
\label{sec:experiments}
\subsection{Implementation Details}

To verify the performance of our optimized DeiT model on BPU, we conduct experiments on two datasets: ImageNet~\cite{deng2009imagenet} and Flower Classification~\cite{2020mmclassification}. ImageNet is a large-scale dataset of labeled images, widely used in computer vision research, containing over 14 million images organized into more than 20,000 categories based on the WordNet hierarchy. The Flower Classification dataset is relatively small, containing 2,939 training images and 731 test images across 5 classes: daisy, dandelion, roses, sunflowers, and tulips.

Since ImageNet is widely used in computer vision research, the original DeiT is no exception. Therefore, for ImageNet, we simply utilize the pre-trained model weights provided by the original DeiT paper. To maintain consistency with previous work, we evaluate our models on the ImageNet validation set, which consists of 50,000 images.

As there are no pre-trained DeiT weights for the Flower Classification dataset, we fine-tune the original DeiT model. The fine-tuning configuration is as follows: we use a single RTX 4090 GPU and the AdamW optimizer ($lr=5e\text{-}4 \times 16/64$, $weight\_decay=0.05$, $betas=(0.9, 0.999)$). The learning rate schedule follows a cosine annealing strategy with linear warm-up for 3 epochs, where the warm-up learning rate starts from 0.001 times the base learning rate, and the minimum learning rate is set to 0.01 times the maximum. The model is initialized with pre-trained weights from the original DeiT paper, and backbone layers are frozen during fine-tuning. Training is performed using label smoothing loss (smoothing factor $\epsilon=0.1$) for 100 epochs. Finally, we quantize the obtained weights for performance comparison.

All performance tests are conducted on the Horizon Brain Processing Unit (BPU). For non-quantized models, inference is performed using ONNX Runtime, while quantized models are run using the BPU-specific inference engine.

\subsection{Model Quantization and Calibration Dataset}
After importing the model weights into our proposed BPU Vision Transformer architecture, we first export the model to ONNX format. Since the quantization process relies on a calibration set to determine various parameters, a calibration set consisting of approximately 100 images is essential for our model quantization. To ensure accurate quantization parameter selection, we construct the calibration dataset by randomly sampling from the ImageNet training set. This calibration data plays a vital role in determining optimal quantization parameters, as it helps capture the actual data distribution the model will encounter during inference.

We then import both the ONNX model file and calibration dataset into the Horizon BPU quantization toolchain to perform model quantization and acceleration. Using a min-max quantization strategy, the toolchain automatically converts the floating-point model to INT8 precision while optimizing for the BPU's hardware characteristics.

\subsection{Comparison with Original Model}

\textbf{On ImageNet.}
For a comprehensive evaluation of our quantized DeiT model, we test model accuracy and computational efficiency on the ImageNet validation set. We strictly follow the evaluation methodology established in the original DeiT paper. As explained earlier, due to the restrictions of embedded computing hardware, we evaluate the Tiny, Small, and Base versions of DeiT as well as their distilled counterparts.

\begin{table}[h]
    \caption{Performance Comparison of DeiT Models Before and After Quantization. \emph{Dist.} represents the distilled version of the corresponding model size.}
    \label{tab:model_comparison}
    \centering
    \begin{tabular}{|l|c|c|c|c|}
    \hline
    \multirow{2}{*}{\textbf{Model}} & \multicolumn{2}{c|}{\textbf{Original}} & \multicolumn{2}{c|}{\textbf{After Quantization}} \\
    \cline{2-5}
     & \textbf{Top-1 (\%)} & \textbf{Top-5 (\%)} & \textbf{Top-1 (\%)} & \textbf{Top-5 (\%)} \\
    \hline
    DeiT-Tiny & 72.1 & 91.1 & 70.4 (-1.7) & 90.2 (-0.9) \\
    DeiT-Small & 79.8 & 95.0 & 72.9 (-6.9) & 91.2 (-3.8) \\
    DeiT-Base & 81.8 & 95.6 & 80.4 (-1.4) & 94.6 (-1.0) \\
    \hline
    DeiT-Tiny Dist. & 74.5 & 91.9 & 74.1 (-0.4) & 91.5 (-0.4) \\
    DeiT-Small Dist. & 81.2 & 95.4 & 80.1 (-1.1) & 94.8 (-0.6) \\
    DeiT-Base Dist. & 83.4 & 96.5 & 81.7 (-1.7) & 95.7 (-0.8) \\
    \hline
    \end{tabular}
\end{table}

Table~\ref{tab:model_comparison} presents a performance comparison of DeiT (Data-efficient Image Transformers) models before and after quantization. For each model size (Tiny, Small, and Base), the table shows Top-1 and Top-5 classification accuracies.

The non-quantized models demonstrate superior performance, with accuracy progressively improving from DeiT-Tiny to DeiT-Base. Notably, the distilled variants consistently outperform their counterparts, with DeiT-Base Dist. achieving the highest Top-1 accuracy of $83.4\%$ and Top-5 accuracy of $96.5\%$.

After quantization, all models experience some performance degradation in Top-1 accuracy, with most models maintaining accuracy within $2\%$ of their original performance, demonstrating an effective balance between computational efficiency and model quality. The only exception is DeiT-Small, which experiences a notable accuracy drop of $6.9\%$ during quantization. A possible reason is that compared with DeiT-Tiny (5M parameters), DeiT-Small (22M parameters) achieves a relatively large improvement, close to that of DeiT-Base (86M parameters). Therefore, DeiT-Small might not be as robust as other models under quantization. The Top-5 accuracy degradation is much lower than Top-1, with respective drops of $0.9\%$, $3.8\%$, and $0.9\%$ for DeiT-Tiny, DeiT-Small, and DeiT-Base, suggesting that the quantized models maintain their ability to rank the correct class among the top predictions, even when making errors in their Top-1 predictions. This demonstrates the performance potential of the quantized model.

The distilled models exhibit remarkable resilience to quantization, with minor accuracy reductions. For instance, the quantized DeiT-Base Dist. maintains a strong Top-1 accuracy of $81.7\%$, with an accuracy reduction of only $1.7\%$. This robustness is particularly evident in distilled models, where the accuracy degradation remains consistently low (ranging from $0.4\%$ to $1.7\%$). In addition, the Top-5 accuracy degradation ($0.4\%-0.8\%$) is consistently smaller than the Top-1 degradation. These results fully demonstrate that the quantized versions successfully retain the performance of the original models, also reflecting the effectiveness of our optimization of the DeiT model.

\textbf{On Flower Classification.} The experimental results on the Flower Classification dataset are presented in Table~\ref{tab:flower_accuracy}. All three DeiT variants achieve high accuracy above 96\% with full precision weights. After quantization, the models maintain strong performance with minimal accuracy degradation. Notably, DeiT-Small exhibits the best robustness to quantization with only a 0.3\% accuracy drop, while preserving the highest accuracy of 97.4\% among all quantized models. In contrast, DeiT-Tiny shows a relatively larger performance degradation of 1.4\%, though still maintaining acceptable accuracy above 95\%. The possible cause for such results is similar to that observed on ImageNet. Since the Flower Classification dataset is relatively small, the Tiny version of DeiT can achieve accuracy comparable to larger models, but may not be as robust. These results demonstrate that our quantization approach successfully preserves the high classification performance of DeiT models while enabling deployment on resource-constrained devices.

\begin{table}[h]
  \caption{Classification Accuracy on Flower Dataset Before and After Quantization}
  \label{tab:flower_accuracy}
  \centering
  \begin{tabular}{|l|c|c|c|}
  \hline
  \multirow{2}{*}{\textbf{Model}} & \textbf{Full Precision} & \textbf{Quantized} & \textbf{Accuracy Drop} \\
   & \textbf{(\%)} & \textbf{(\%)} & \textbf{(\%)} \\
  \hline
  DeiT-Tiny & 97.0 & 95.6 & 1.4 \\
  DeiT-Small & 97.7 & 97.4 & 0.3 \\
  DeiT-Base & 97.5 & 97.0 & 0.5 \\ 
  \hline
  \end{tabular}
\end{table}

\textbf{Efficiency Tests.} Another criterion is the speedup achieved by our optimized model after quantization. Table~\ref{tab:inference_time} illustrates the inference time for various DeiT model variants and the speedup achieved after quantization. It can be observed that, for the original DeiT models, inference time is proportional to model size, and all models achieve performance gains after quantization.

\begin{table}[h]
    \caption{Inference Time Comparison for Different Model Configurations}
    \label{tab:inference_time}
    \centering
    \begin{tabular}{|l|c|c|}
    \hline
    \textbf{Model Type} & \textbf{Inference Time (s)} & \textbf{Speedup Factor} \\
    \hline
    Tiny & $0.61$ & $1.00$ \\
    Small & $2.05$ & $1.00$ \\
    Base & $7.41$ & $1.00$ \\         
    \hline
    Tiny (Quantized) & $0.47$ & $1.30$ \\
    Small (Quantized) & $0.95$ & $2.16$ \\
    Base (Quantized) & $1.94$ & $3.82$ \\
    \hline
    \end{tabular}
\end{table}

Most notably, the Base model demonstrates a significant $3.82\times$ speedup in inference time, while the Small and Tiny variants achieve speedups of $2.16\times$ and $1.30\times$, respectively. This indicates that quantization benefits are more pronounced in larger model architectures and implies further performance potential for even larger models.

\subsection{Model Predictions on Annotation Errors}
During our evaluation, we observed DeiT and our quantized model's behavior on the ImageNet validation set. While examining the mismatches between model predictions and ground truth labels, we discovered that DeiT occasionally generates predictions that appear more accurate than the official ImageNet annotations.

\begin{figure}[!h]
  \centering
  \includegraphics[width=0.48\textwidth]{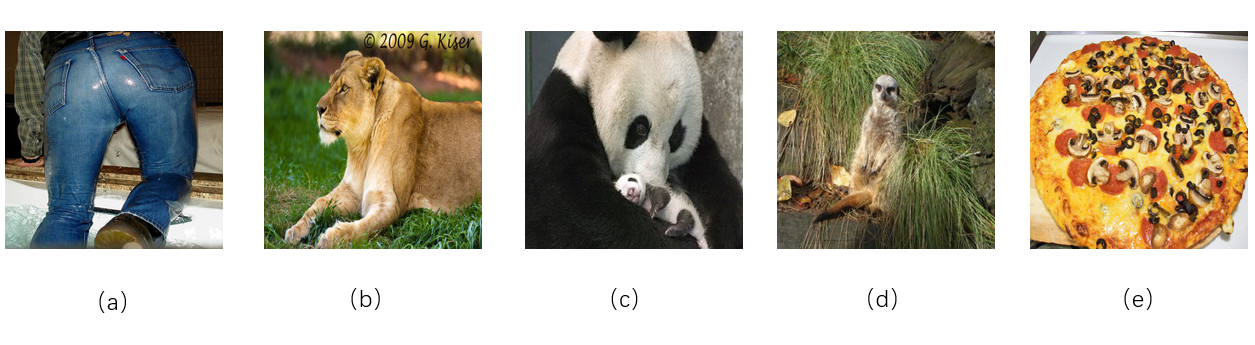}
  \caption{Examples of misclassified images from the validation set. These samples demonstrate typical cases where the model made incorrect predictions, providing insights into the model's limitations and potential areas for improvement.}
  \label{fig:misclassified}
\end{figure}

Specifically, we identified several cases where images were evidently mislabeled in the validation set, yet DeiT produced semantically correct classifications. These images are shown in Figure~\ref{fig:misclassified}.

From the results shown in Table~\ref{tab:predictions}, we can observe that both DeiT and our quantized model make consistent predictions across all cases. These findings suggest that the actual accuracy of both models may be higher than the reported values, as the validation set contains annotation errors. This also indicates that DeiT's accuracy should be slightly higher than the accuracy reported in the original paper, as demonstrated by its correct predictions on several mislabeled samples.

\begin{table}[!h]
    \centering
    \caption{Analysis of Model Predictions on Images with Incorrect Annotations}
    \begin{tabular}{|c|c|c|c|}
    \hline
    Image & Incorrect Annotation & DeiT Prediction & Quantized Prediction \\
    \hline
    (a) & tub & jeans & jeans \\
    (b) & patas monkey & lion & lion \\
    (c) & red panda & giant panda & giant panda \\
    (d) & red panda & meerkat & meerkat \\
    (e) & dough & pizza & pizza \\
    \hline
    \end{tabular}
    \label{tab:predictions}
\end{table}

\section{CONCLUSIONS}
\label{sec:conclusion}
In this paper, we proposed a specialized DeiT-based Vision Transformer model for the BPU platform, representing the first successful deployment of Vision Transformers on BPU to fully exploit its acceleration capabilities. Our model effectively leverages the BPU's hardware advantages for INT8 precision convolution operations. Experimental results demonstrate that our approach achieves competitive performance in both efficiency and accuracy. Through extensive testing, we observed that the quantized model maintains high prediction accuracy while delivering significant speedup on BPU hardware, making it a practical solution for real-world deployment scenarios.

\addtolength{\textheight}{-12cm}   



\bibliographystyle{IEEEtran}  
\bibliography{refs}           

\end{document}